\documentclass[preprint,12pt]{elsarticle}

\usepackage{amssymb}
\usepackage{amsmath}
\usepackage[ruled,linesnumbered]{algorithm2e}
\usepackage{makecell}
\usepackage{multirow}
\usepackage{booktabs}
\usepackage{tabularx}
\usepackage{graphicx}
\usepackage[table]{xcolor}

\usepackage{subfigure}

\journal{{Pattern Recognition}}

\begin{document}

\begin{frontmatter}



\title{DQE-CIR: Distinctive Query Embeddings through Learnable Attribute Weights and Target Relative Negative Sampling in Composed Image Retrieval}


\author{Geon Park} 
\author{Ji-Hoon Park}
\author{Seong-Whan Lee}
\affiliation{organization={Department of Artificial Intelligence, Korea University},
            city={Seoul},
            postcode={02841}, 
            country={Republic of Korea}}

\begin{abstract}
Composed image retrieval (CIR) addresses the task of retrieving a target image by jointly interpreting a reference image and a modification text that specifies the intended change. Most existing methods are still built upon contrastive learning frameworks that treat the ground truth image as the only positive instance and all remaining images as negatives. This strategy inevitably introduces relevance suppression, where semantically related yet valid images are incorrectly pushed away, and semantic confusion, where different modification intents collapse into overlapping regions of the embedding space. As a result, the learned query representations often lack discriminativeness, particularly at fine-grained attribute modifications.
To overcome these limitations, we propose distinctive query embeddings through learnable attribute weights and target relative negative sampling (DQE-CIR), a method designed to learn distinctive query embeddings by explicitly modeling target relative relevance during training. DQE-CIR incorporates learnable attribute weighting to emphasize distinctive visual features conditioned on the modification text, enabling more precise feature alignment between language and vision. Furthermore, we introduce target relative negative sampling, which constructs a target relative similarity distribution and selects informative negatives from a mid-zone region that excludes both easy negatives and ambiguous false negatives. This strategy enables more reliable retrieval for fine-grained attribute changes by improving query discriminativeness and reducing confusion caused by semantically similar but irrelevant candidates.
We conduct extensive experiments on standard CIR benchmarks, including FashionIQ and CIRR, under both supervised and zero-shot retrieval settings. DQE-CIR consistently achieves superior performance across multiple evaluation metrics, demonstrating improvements in both coarse retrieval accuracy and fine-grained semantic alignment compared to existing methods. Detailed ablation studies and qualitative analyses further validate the contribution of each component and confirm that modeling target relative negative samples is crucial for robust and reliable CIR.
\end{abstract}


\begin{highlights}
\item We propose a composed image retrieval method that learns more distinctive query representations for fine-grained image retrieval.
\item We introduce a target relative negative sampling strategy that selects informative negatives while avoiding overly easy or false negatives.
\item We design a pairwise learning objective that clearly separates images matching the intended modification from confusing candidates.
\item We show that the proposed components consistently improve retrieval accuracy on multiple composed image retrieval benchmarks.
\end{highlights}


\end{frontmatter}



\section{Introduction}
Composed Image Retrieval (CIR) aims to retrieve images that apply the modification specified by the user based on a reference image and a textual description. Unlike existing text-to-image or image-to-image retrieval methods~\cite{YANG2024110273, 10.1145/3581783.3611817, ZHAO2025111696}, CIR enables fine-grained and controllable retrieval by jointly modeling visual context and linguistic modification, making it particularly suitable for interactive retrieval where users iteratively refine their intent. Recent advances in CIR~\cite{li2025learning, Huynh_2025_CVPR} have significantly expanded its practical impact across real-world applications such as fashion search, product recommendation, and content management systems, where precise attribute changes such as color, shape, quantity, or appearance must be accurately applied in the retrieved results~\cite{du2025survey, song2025comprehensive}. As a result, recent research has increasingly focused on developing distinctive query representations that faithfully capture subtle semantic shifts while remaining robust to visually similar but irrelevant candidates, exploring directions such as improved multimodal fusion and attribute-aware modeling~\cite{wan2024cross, zhang2024multimodal}. Although extensive research has improved the overall capability of CIR methods, most existing methods~\cite{baldrati2023composed, yang2024decomposing, KE2025111096} still rely on contrastive learning, where the target image is treated as the only positive object, and all remaining images are classified as negatives. This introduces two limitations that arise in CIR.

\begin{figure*}
    \centering
    \includegraphics[width=\textwidth]{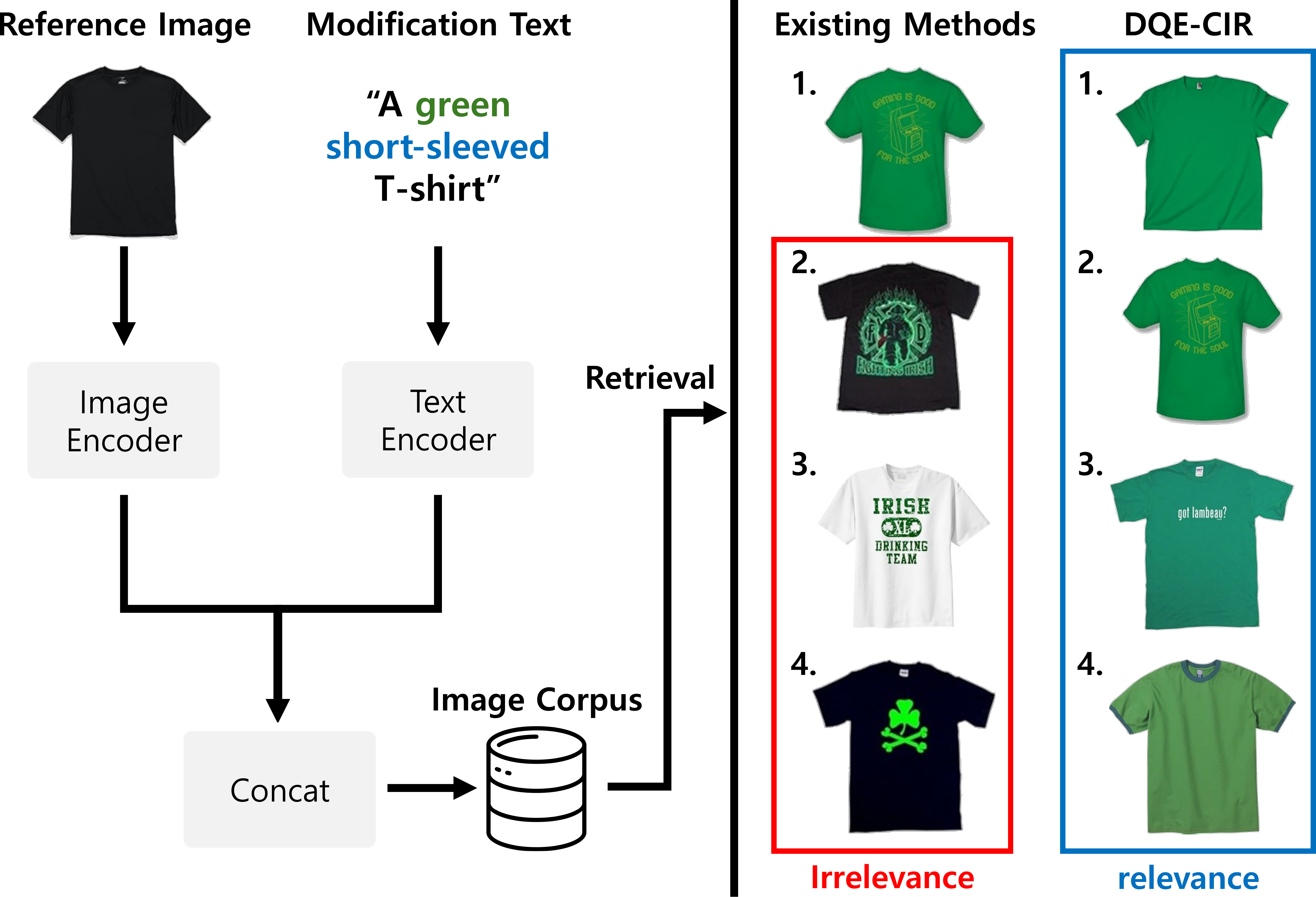}
    \caption{
        Example of attribute-aware relevance separation in CIR. 
        Given a reference image and the modification text specifying green and short-sleeved, candidate images are grouped according to their attribute alignment. The red box shows irrelevant images that violate key attributes and commonly appear in top-ranked results of existing CIR models. In contrast, the blue box shows relevant images that satisfy both attributes—results that DQE-CIR retrieves consistently through attribute-aware query embeddings and distinctive ranking constraints.
    }
    \label{fig1}
\end{figure*}

As illustrated in Fig.~\ref{fig1}, the first limitation is relevance suppression. Images that match modified attributes of the user (i.e., “green” and “short-sleeved”) are often treated as negatives simply because they are not labeled as the target. These relevant images share essential attributes with the target but are gradually penalized during training. Consequently, they receive lower similarity scores and can be ranked below irrelevant images that do not satisfy the conditions of the user. Since real-world users typically expect multiple relevant results rather than a single exact match, the suppression of such images directly undermines retrieval quality and user satisfaction~\cite{10.1145/3723879, Tian_2025_CVPR}.
The second limitation visible in Fig.~\ref{fig1} is semantic confusion. Contrastive learning does not explicitly model attribute discriminativeness, which embeddings for different queries tend to collapse into nearby regions of the representation space. As a result, the model struggles to separate queries with different attribute modifications, leading to reduced discriminativeness when subtle attributes such as color or sleeve length must be distinguished~\cite{10.1145/3664647.3680808}.

To address these limitations, we propose Distinctive Query Embeddings for Composed Image Retrieval through Learnable Attribute Weights and Target Relative Negative Sampling (DQE-CIR), a framework designed to produce more distinctive and attribute-aware query embeddings while mitigating performance degradation caused by false negatives. The first component of DQE-CIR is Learnable Attribute Weights, which enable the model to adaptively regulate the contribution of key attributes such as color and shape within the composed query representation. By assigning learnable importance to attribute-specific features derived from textual modifications, the model emphasizes attributes that are critical for a given retrieval intent while suppressing less informative features. This adaptive weighting leads to a more distinctive embedding space, in which queries applying different attribute changes are better distinguished from one another, thereby facilitating more reliable alignment with target images in CIR.

The second component is Target Relative Negative Sampling (TRNS). Instead of treating all non-target images as negatives, TRNS evaluates each candidate relative to the target by computing a $\Delta$-score and identifying a mid-zone region that contains images sufficiently similar to provide meaningful supervision but not so similar that they constitute false negatives. This target relative perspective prevents both relevance suppression and overly easy negative sampling, allowing the model to focus training on semantically informative samples. After defining this target relative negative set, Pairwise Learning is applied. Unlike standard contrastive learning, which compares a positive sample against many negatives simultaneously, pairwise learning focuses on a single selected negative for each query. This strengthens the ranking margin between relevant and less relevant images, allowing the model to train clearer preference ordering and to form a distinctive embedding space.

We conduct extensive experiments on standard CIR benchmarks, including FashionIQ and CIRR, under both supervised and zero-shot retrieval settings. Quantitative results evaluated using Recall@K and mean Average Precision (mAP) demonstrate that DQE-CIR consistently outperforms existing methods, achieving improvements in both overall retrieval accuracy and fine-grained retrieval performance. Qualitative evaluations further show that the proposed method retrieves images that better apply the intended modifications, producing more accurate and semantically appropriate retrieval results compared to prior methods. In addition, ablation studies systematically verify the contribution of each component, confirming that utilizing target relative negative samples and applying attribute emphasis pairwise learning are crucial for reliable CIR.

The main contributions of this paper are summarized as follows: 
\begin{itemize} 
    \item We introduce DQE-CIR, a framework that constructs distinctive and attribute-sensitive query embeddings through learnable attribute weights and auxiliary attribute queries.
    \item We present a Target Relative Negative Sampling, a negative selection strategy that identifies a mid-zone of target relative negatives using a $\Delta$-score band. This mid-zone excludes false negatives and removes excessively easy negatives, enabling training to focus on semantically informative and properly challenging samples.
    \item We demonstrate the effectiveness of DQE-CIR through extensive experiments, showing that it outperforms existing methods in various benchmarks. The improvements are particularly pronounced in mitigating relevance suppression and semantic confusion, two limitations of existing CIR methods.
\end{itemize}

\section{Related Work}
\subsection{Vision-Language Models}
Vision–Language Models seek to learn joint representations from paired visual and textual data, enabling a single model to reason across modalities and support a wide range of downstream tasks such as retrieval, captioning, and multimodal understanding~\cite{zhou2022conditional, WANG2025111085}. Foundational dual-encoder methods~\cite{dai2023instructblip, lin2024vila, CaoJHZLWC024}, exemplified by CLIP~\cite{pmlr-v139-radford21a}, align global image and text embeddings through large-scale contrastive pretraining, demonstrating strong generalization and zero-shot capabilities. Subsequent architectures, such as BLIP~\cite{li2022blip}, introduce cross-modal fusion strategies that integrate linguistic context into visual feature processing, enhancing the model capacity for fine-grained semantic reasoning~\cite{li-etal-2023-lavis, JIANG2025111144}. A particularly influential advancement is BLIP-2~\cite{li2023blip}, which adopts a modular design with a frozen vision encoder and a frozen large language model connected by a lightweight Q-Former~\cite{kim-etal-2024-towards-efficient, Azad_2025_CVPR, Tang_Huang_Zheng_2025}. The Q-Former employs trainable query tokens that attend to the frozen visual backbone and extract task-relevant features, which are then projected into the language model space through cross-attention. This strategy not only yields expressive multimodal representations but also significantly reduces training cost by limiting the number of trainable parameters~\cite{LI2025111088, Azad_2025_ICCV}.
In DQE-CIR, we build upon BLIP-2 as the underlying vision–language backbone. Specifically, we utilize its pretrained image encoder, text encoder, and Q-Former to construct the initial multimodal embedding space for composed queries and candidate images. Unlike methods that use BLIP-2 solely for feature extraction, DQE-CIR fine-tunes the Q-Former and projection layers to generate distinctive and composition-aware query embeddings that are sensitive to attribute variations expressed in textual modifications. This adaptation enables the model to leverage strong pretrained priors from BLIP-2 while adapting its representations to the specific demands of CIR.

\subsection{Composed image retrieval}
Composed image retrieval (CIR) addresses the task of retrieving a target image from a database given a multimodal query that combines a reference image with a modification text \cite{feng2024improving, baldrati2022effective}. Recent methods~\cite{sun2025leveraging, zhu2025interactive,2024visionbylanguage} build on large vision–language backbones and differ mainly in how they form and train the composed query representation. CLIP4Cir~\cite{baldrati2023composed} fine-tunes CLIP and introduces a lightweight composition network that fuses the reference image feature and the relative caption into a single query embedding, which is then optimized with triplet-style contrastive losses for CIR benchmarks. Extending this method, BLIP4CIR~\cite{liu2024bi} leverages BLIP-style representations and bi-directional training, where the model is trained not only to retrieve the target image given a composed query but also to retrieve the reference image given the target and the modification text, thereby improving the robustness of the composition function and better exploiting the symmetry in CIR. Beyond small- to medium-scale curated datasets, CoVR~\cite{ventura2024covr} focuses on composed video retrieval by exploiting web-scale video captions, automatically constructing large-scale video–text–video training pairs, and adapting a BLIP-2-based model to capture compositional modifications in video retrieval. More recently, SPRC~\cite{bai2024sentencelevel} proposes to learn sentence-level prompts appended to the modification text, using pretrained vision–language models such as BLIP-2 and a combination of image–text contrastive loss and prompt-alignment loss, so that standard text-based retrieval models can be reused for CIR without designing task-specific composition networks. Complementary to prompt-based methods, QuRe~\cite{kwak2025qure} introduces a hard negative sampling strategy that explicitly accounts for the relevance of non-target images, constructing query-relevant negatives to encourage the model to distinguish between varying degrees of semantic similarity within the candidate set. Despite these advances, most prior CIR methods~\cite{li2024cross, xing2025context, 10888719} still rely on contrastive learning that treats the ground-truth image as the only positive and all remaining images as negatives. This method often causes relevance suppression, where semantically relevant images are incorrectly pushed away as false negatives, and semantic confusion, where distinct modification intents collapse into overlapping regions in the embedding space~\cite{suo2024knowledge, Kolouju_2025_CVPR}. 
In contrast, DQE-CIR adopts a pairwise learning paradigm over composed query–image pairs, directly optimizing relative preference scores so that more relevant images are consistently ranked above less relevant ones, thereby mitigating Relevance Suppression and Semantic Confusion while producing a more distinctive and composition-aware retrieval space.

\section{Method}
\begin{figure*}
    \centering
    \includegraphics[width=\textwidth]{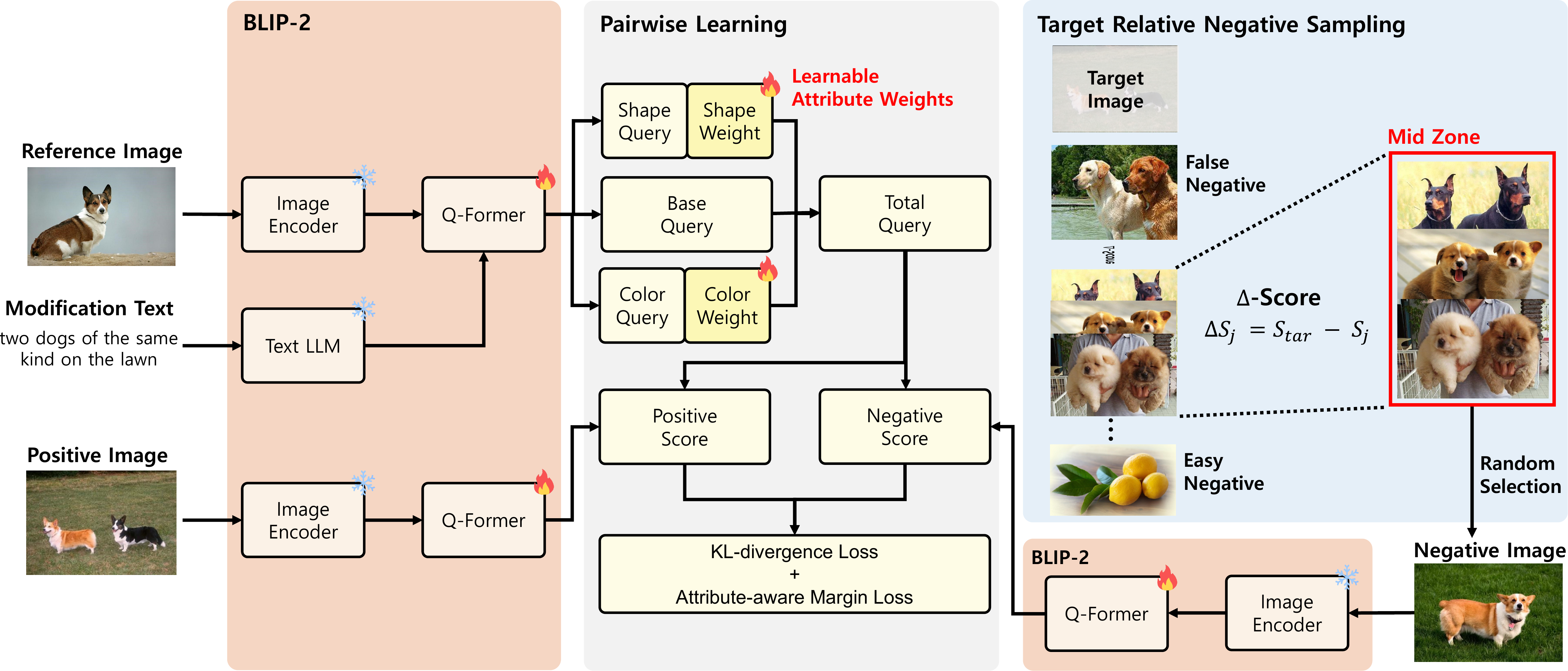}
    \caption{
    Overview of the proposed DQE-CIR framework.
    DQE-CIR first encodes the reference image, modification text, and candidate images using BLIP-2 to obtain base query representations. Learnable Attribute Weights then enhance the color- and shape-specific sub-queries, which are combined to form the final attribute-aware query embedding and optimized through KL-divergence and attribute-aware margin losses. In parallel, Target Relative Negative Sampling (TRNS) selects a single negative from the $\Delta$-score–based mid-zone, enabling distinctive pairwise ranking training. Together, these components strengthen fine-grained attribute sensitivity and improve retrieval discriminativeness.
    }
    \label{fig2}
\end{figure*}
To enhance distinctive query embeddings and improve retrieval performance in CIR, we propose DQE-CIR, a framework that applies attribute-aware query construction, target relative negative sampling, and single-negative ranking optimization. DQE-CIR aims to generate query representations that more clearly capture the intended attribute modifications, thereby reducing semantic ambiguity and improving fine-grained retrieval accuracy. The overall framework is illustrated in Fig.~\ref{fig2}, which shows how the model integrates learnable attribute weights, dynamic negative selection based on $\Delta$-score bands, and a single-negative pairwise learning strategy to construct a more distinctive and composition-aware embedding space. Through these components, DQE-CIR effectively strengthens attribute sensitivity, mitigates semantic confusion, and enhances the ability of the model to apply the modification text to the reference image in a structurally meaningful way.

\subsection{Problem Definition}
We consider a CIR dataset $D$ composed of triplets. Each triplet is defined as $(x_{\mathrm{ref}}, x_{\mathrm{txt}}, y_{\mathrm{tar}})$, where $x_{\mathrm{ref}}$ denotes the reference image, $x_{\mathrm{txt}}$ specifies the modification text describing the intended change, and $y_{\mathrm{tar}}$ is the target image that accurately applies the intended modification. The image corpus is denoted as $\{I_j\}_{j=1}^{N}$, where $N$ is the total number of candidate images. For each query pair $(x_{\mathrm{ref}}, x_{\mathrm{txt}})$, the goal is to construct a query embedding $q$ that accurately captures the intended transformation from $x_{\mathrm{ref}}$ to $y_{\mathrm{tar}}$. During retrieval, the model computes similarity scores between the query embedding and the representation of every image $I_j \in I$, and ranks them accordingly. The correct target image $y_{\mathrm{tar}}$ should receive the highest similarity score, and all other images should be ordered based on their semantic relevance.

\subsection{Learnable Attribute Weights}
To enhance the ability to apply fine-grained attribute modifications, learnable attribute weights are integrated into the Q-former-based query encoding process. The Q-former receives $x_{\mathrm{ref}}$ and $x_{\mathrm{txt}}$, and produces a set of $U$ query tokens $Q = \{q_u\}_{u=1}^{U}$, where each token $q_u$ corresponds to one of the $U$ latent learnable queries used by the Q-former to extract cross-modal features. A pooled representation of these tokens yields the composed query embedding $q$.
To explicitly emphasize attribute-sensitive features, attribute-specific sub-query features $q_{\mathrm{color}}$ and $q_{\mathrm{shape}}$ are extracted from Q-former attention outputs conditioned on color-related and shape-related terms in the modification text. The attribute-enhanced query embedding $q^{*}$ is defined by the following equation:
\begin{equation}
q^{*} = q + w_{\mathrm{color}} \cdot q_{\mathrm{color}} + w_{\mathrm{shape}} \cdot q_{\mathrm{shape}},
\end{equation}
where $w_{\mathrm{color}}$ and $w_{\mathrm{shape}}$ are learnable scalar weights that modulate the contribution of color- and shape-sensitive features to the final embedding. This formulation emphasizes distinctive attribute dimensions within the embedding space, enabling the composed query representation to align more faithfully with the intended modification while remaining distinguishable from visually similar distractors during retrieval.

\subsection{Target Relative Negative Sampling} \label{sec:Target}
\begin{figure*}
    \centering
    \includegraphics[width=\textwidth]{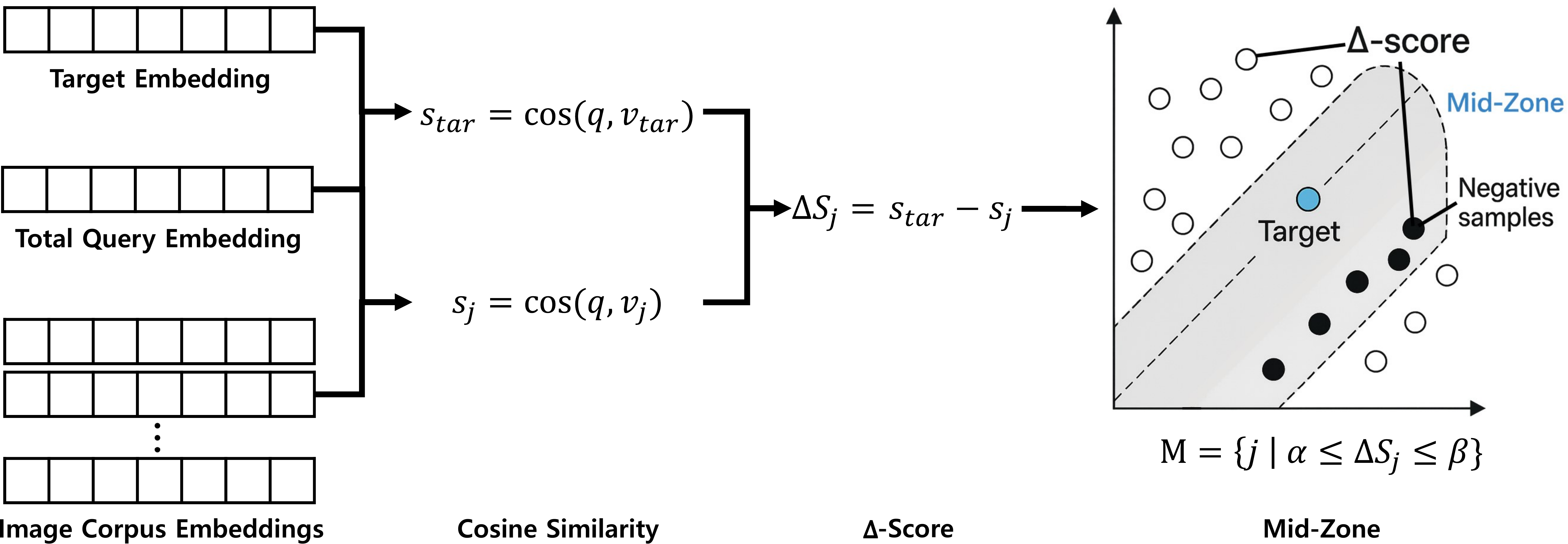}
    \caption{
    Illustration of Target Relative Negative Sampling (TRNS).
    The total query embedding is compared with the target embedding and all image corpus embeddings using cosine similarity to derive relevance scores. Subsequently, the $\Delta$-score is computed as the difference between the target similarity and each candidate similarity. Candidate images whose $\Delta$-score falls within a predefined mid-zone are designated as informative negatives for training.
    }
    \label{fig3}
\end{figure*}
To identify informative negatives that contribute to a more distinctive query embedding, DQE-CIR constructs a target relative $\Delta$-score distribution for each composed query. Given a reference image and modification text, the model first encodes them into a joint query embedding $q$ that represents how the reference image should be applied with the specified attribute changes. In parallel, every image $I_j$ in the image corpus $I$ is encoded into an embedding $v_j$, and the corresponding target image $y_{\mathrm{tar}}$ is encoded into $v_{\mathrm{tar}}$. The target embedding $v_{\mathrm{tar}}$ is treated as the positive ground truth because it is expected to apply the intended modification most accurately. As shown in Fig.~\ref{fig3}, the model computes cosine similarities between the query embedding and all image embeddings:
\begin{equation}
s_j = \cos(q^{*}, v_j), \qquad
s_{\mathrm{tar}} = \cos(q^{*}, v_{\mathrm{tar}}),
\end{equation}
where $s_j$ measures how closely a candidate image $I_j$ applies the intended transformation, and $s_{\mathrm{tar}}$ measures how well the target image applies the same transformation. Using these similarity scores, DQE-CIR defines the target relative $\Delta S$-value for each candidate image as:
\begin{equation}
\Delta S_j = s_{\mathrm{tar}} - s_j,
\end{equation}
where $\Delta S_j$ quantifies the similarity gap between the target image and candidate $I_j$, with smaller values indicating candidates that are closer to the target in the embedding space. This yields a one-dimensional $\Delta S$ distribution centered around the target similarity.

To focus on moderately challenging negatives, DQE-CIR defines a $\Delta S$-based mid-zone by discarding overly easy negatives with very large $\Delta S$-values and overly confusing negatives whose $\Delta S$-values are too close to zero. Formally, the mid-zone is defined as:
\begin{equation}
M = \{ j \mid \alpha \leq \Delta S_j \leq \beta \},
\end{equation}
where $\alpha$ and $\beta$ denote the lower and upper bounds of the $\Delta S$ band, respectively, ensuring that only candidates with intermediate similarity gaps are preserved and later used for single-negative sampling.

\begin{algorithm}[!ht]
\caption{DQE-CIR: Distinctive Query Embeddings through Learnable Attribute Weights and Target Relative Negative Sampling in Composed Image Retrieval}
\label{algo}
\KwIn{
Reference image $x_{\mathrm{ref}}$, modification text $x_{\mathrm{txt}}$, 
target image $y_{\mathrm{tar}}$, image corpus $I=\{I_j\}_{j=1}^{N}$, 
mid-zone bounds $(\alpha,\beta)$, temperature $\tau$, margins $m, m_a$, 
weight $\lambda_{\mathrm{rank}}$
}
\KwOut{Total loss $L_{\mathrm{total}}$}

$q \leftarrow f_{\mathrm{Q}}(x_{\mathrm{ref}}, x_{\mathrm{txt}})$

$q_{\mathrm{color}}, q_{\mathrm{shape}} \leftarrow f_{\mathrm{A}}(x_{\mathrm{txt}})$

$q^{*} \leftarrow q + w_{\mathrm{color}} q_{\mathrm{color}} + w_{\mathrm{shape}} q_{\mathrm{shape}}$

$v_{\mathrm{tar}} \leftarrow f_{\mathrm{I}}(y_{\mathrm{tar}})$

$s_{\mathrm{tar}} \leftarrow \cos(q^{*}, v_{\mathrm{tar}})$

\ForEach{$I_j \in I \setminus \{y_{\mathrm{tar}}\}$}{
$v_j \leftarrow f_{\mathrm{I}}(I_j)$

$s_j \leftarrow \cos(q^{*}, v_j)$

$\Delta S_j \leftarrow s_{\mathrm{tar}} - s_j$
}

$\mathcal{M} \leftarrow \{ j \mid \alpha \le \Delta S_j \le \beta \}$

$j^{-} \sim \mathcal{M}$ \tcp*{sample a single negative in the mid-zone}

$v_{j^{-}} \leftarrow f_{\mathrm{I}}(I_{j^{-}})$

$s_{j^{-}} \leftarrow \cos(q^{*}, v_{j^{-}})$

$p_{\mathrm{pred}} \leftarrow \mathrm{softmax}\!\left(\frac{[s_{\mathrm{tar}}, \{s_j\}]}{\tau}\right)$

$p_{\mathrm{tar}} \leftarrow [1,0,\dots,0]$

$L_{\mathrm{KL}} \leftarrow \mathrm{KL}(p_{\mathrm{tar}} \,\|\, p_{\mathrm{pred}})$

$L_{\mathrm{main}} \leftarrow \max(0, m - s_{\mathrm{tar}} + s_{j^{-}})$

\ForEach{$a \in \{\mathrm{color}, \mathrm{shape}\}$}{
$s^{(a)}_{\mathrm{tar}} \leftarrow \cos(q_a, v_{\mathrm{tar}})$

$s^{(a)}_{\mathrm{neg}} \leftarrow \cos(q_a, v_{j^{-}})$

$L_a \leftarrow \max(0, m_a - s^{(a)}_{\mathrm{tar}} + s^{(a)}_{\mathrm{neg}})$
}

$L_{\mathrm{total}} \leftarrow L_{\mathrm{KL}} 
+ \lambda_{\mathrm{rank}} L_{\mathrm{main}}
+ w_{\mathrm{color}} L_{\mathrm{color}}
+ w_{\mathrm{shape}} L_{\mathrm{shape}}$

\Return{$L_{\mathrm{total}}$}

\end{algorithm}
\subsection{Learning with Single-Negative Ranking Pairs}
After constructing the mid-zone $M$ from the $\Delta S$ distribution as described in Section.~\ref{sec:Target}, DQE-CIR forms a single target negative ranking pair by randomly selecting one candidate image within $M$. The sampled negative serves as the contrastive counterpart to the correct target image, providing a compact ranking structure that enhances the ability to distinguish fine-grained and attribute-sensitive modifications. Beyond the main composed query, DQE-CIR further introduces color- and shape-specific sub-queries, each designed to highlight attribute-related features that are often underrepresented in existing contrastive learning. These sub-queries supply auxiliary pairwise constraints that operate jointly with the primary divergence objective. The KL divergence loss for the composed query is defined as:
\begin{equation}
L_{\mathrm{KL}} = KL(p_{\mathrm{tar}} \,\|\, p_{\mathrm{pred}}),
\end{equation}
where $p_{\mathrm{tar}}$ denotes the target distribution over the image corpus associated with the reference--text composition, and $p_{\mathrm{pred}}$ represents the predicted distribution over candidate image representations. 
To enforce a ranking margin between the target and the mid-zone negative, the main pairwise ranking loss is defined as:
\begin{equation}
L_{\mathrm{main}} = \max(0,\, m - s_{\mathrm{tar}} + s_{\mathrm{neg}}),
\end{equation}
where $m$ is a margin hyperparameter, $s_{\mathrm{tar}}$ denotes the similarity between the composed query embedding and the target image, and $s_{\mathrm{neg}}$ represents the similarity with the sampled negative.
For each attribute-aware sub-query $q^{(a)}$, where $a \in \{\mathrm{color}, \mathrm{shape}\}$, the auxiliary margin loss is defined as:
\begin{equation}
L_{a} = \max(0,\, m_{a} - s^{(a)}_{\mathrm{tar}} + s^{(a)}_{\mathrm{neg}}),
\end{equation}
where $s^{(a)}_{\mathrm{tar}}$ and $s^{(a)}_{\mathrm{neg}}$ denote the similarities computed using the attribute-specific sub-query, and $m_{a}$ controls the margin strength for each attribute.
The final training objective that combines KL divergence, the main ranking constraint, and attribute-aware auxiliary losses is defined as:
\begin{equation}
L_{\mathrm{total}} = L_{\mathrm{KL}} + \lambda_{\mathrm{rank}} L_{\mathrm{main}}
+ w_{\mathrm{color}} L_{\mathrm{color}}
+ w_{\mathrm{shape}} L_{\mathrm{shape}},
\end{equation}
where $\lambda_{\mathrm{rank}}$ regulates the influence of the main ranking loss, and $w_{\mathrm{color}}$ and $w_{\mathrm{shape}}$ are learnable scalar weights that adaptively modulate the influence of color- and shape-specific features in both query representation and attribute-aware supervision during training. The pseudocode is presented in Algorithm~\ref{algo}, which summarizes the overall learning procedure of DQE-CIR by integrating attribute-aware query construction, target relative negative sampling, and single-negative pairwise optimization into a unified training method. Given a reference image and a modification text, the method first constructs a composed query embedding that adaptively emphasizes distinctive attribute features through learnable attribute weights, enabling precise alignment between textual modifications and visual content. Instead of treating all non-target images as negatives, DQE-CIR leverages target relative similarity differences to identify informative negatives and selects a single representative negative for training, which provides a clear ranking. This design avoids relevance suppression caused by existing contrastive learning and reduces semantic confusion among visually similar candidates.

\section{Experiment}
\subsection{Experimental Setting}
\paragraph{Evaluation Datasets}
We follow two standard benchmarks for CIR, FashionIQ~\cite{wu2021fashion} and CIRR~\cite{liu2021image}. FashionIQ contains fashion product images paired with modification texts that apply fine-grained attribute changes such as color, material, and sleeve length. This dataset evaluates how effectively a retrieval system applies subtle attribute variations while maintaining the identity of the reference item. CIRR covers more diverse real-world scenes in which the modification text applies broader attribute and shape-oriented changes under larger visual variation. Its subset-based evaluation protocol further examines the ability to distinguish visually similar candidates. Together, FashionIQ and CIRR provide complementary perspectives for assessing whether a method constructs distinctive query embeddings and applies attribute-aware modifications across object-centric and scene-centric retrieval settings.

\paragraph{Implementation Details}
We conduct all experiments on a single NVIDIA A6000 GPU with 48 GB memory using a unified training pipeline built upon a BLIP-2 backbone that applies reference images, target images, and composed queries to a shared embedding space. We train the model for 50 epochs on CIRR and 30 epochs on FashionIQ with a batch size of 128 and adopt mixed precision training for computational efficiency. We optimize all parameters using AdamW~\cite{loshchilov2018decoupled} with weight decay, and we apply a cosine learning rate schedule that gradually reduces the step size across epochs. The training objective integrates a KL divergence term that encourages the composed query to apply the target image more strongly than any candidate, together with a margin-based ranking loss that maintains a similarity margin between the target and the hardest negative sample. We further incorporate color-focused and shape-focused auxiliary queries, whose contributions are controlled by learnable non-negative weights updated jointly with the primary parameters. During training, the target relative negative set is redefined following a fixed interval-based schedule. We first employ an initial warm-up, during which negative samples are not selectively refined and instead consist of the entire image corpus except for the target image, allowing the model to establish a stable similarity distribution. After the warm-up phase, the training process is divided into 5 equal-length intervals, and the target relative negative set is updated once at the beginning of each interval based on the current model predictions. The selected negatives are then kept fixed and reused throughout the remainder of the corresponding interval, without further refinement. During this training process, the target relative negative set is refreshed a total of 5 times over the course of training. This interval-based update strategy balances training stability and computational efficiency, while ensuring that the selected target relative negatives remain informative as the embedding space evolves.

\subsection{Evaluation Metrics}
We evaluate retrieval performance using three metrics that jointly measure how effectively a composed query applies the intended modification and retrieves the correct target image from a large candidate set. Recall@K quantifies the proportion of queries for which the target image is ranked within the top K retrieved results, providing a direct indicator of whether the similarity scores induced by the composed query embedding successfully prioritize the correct target over non-target candidates. Recall@K implies that the method applies the modification text to the reference image in a way that yields a distinctive ordering of images in the corpus, especially around the highest ranks, where retrieval quality is most critical. 

On the CIRR dataset, the evaluation additionally includes Recall$_{subset}$@K, which is computed in a restricted subset of images selected to be semantically and visually similar to the reference image, often sharing global content and local features. This metric focuses on the most challenging cases in which subtle attribute differences, such as color or shape, determine the correct answer, and it evaluates whether the method can apply fine-grained modifications while maintaining clear separation from visually similar distractors. In addition, we report the Average metric, which is defined as the mean of Recall@K over several K values specified by each benchmark protocol, which summarizes retrieval stability across different ranking depths rather than at a single setting. Considering Recall@K, Recall$_{subset}$@K, and Average together enables a comprehensive assessment that covers both overall retrieval accuracy and attribute-aware robustness, and reveals how consistently the proposed method constructs distinctive, modification-aligned query embeddings under varying difficulty levels and corpus conditions.

\subsection{Quantitative Evaluation}
We quantitatively compare our method with a diverse set of existing CIR methods to assess how effectively each system applies the modified text to the reference image and constructs distinctive query embeddings for ranking. This comparison spans both object-centric and scene-centric benchmarks, enabling a comprehensive evaluation of retrieval accuracy under varying levels of visual complexity. For each dataset, we report performance using multiple Recall@K metrics, along with subset-based measures. This enables an evaluation of global retrieval effectiveness over the full image corpus as well as fine-grained target discriminativeness under controlled subset-level retrieval settings. By situating our DQE-CIR framework alongside existing CIR methods, we evaluate whether our attribute-aware weighting and TRNS-driven negative sampling strategy contribute meaningfully to constructing more distinctive and modification-aligned query embeddings.
\begin{table*}[t!]
\caption{Retrieval quantitative performance of diverse CIR models on the FashionIQ validation dataset. Results are reported for the Dress, Shirt, and Toptee categories using Recall@10 and Recall@50. Bold indicates the best performance for each metric.}
\vspace{3pt}
\centering
\resizebox{\textwidth}{!}{%
\begin{tabular}{lcc|cc|cc|cc}
\toprule
\multirow{2}{*}{\textbf{Method}} & \multicolumn{2}{c}{\textbf{Dress}} & \multicolumn{2}{c}{\textbf{Shirt}} & \multicolumn{2}{c}{\textbf{Toptee}} & \multicolumn{2}{c}{\textbf{Average}} \\
\cmidrule(lr){2-3} \cmidrule(lr){4-5} \cmidrule(lr){6-7} \cmidrule(lr){8-9}
 & R@10 & R@50 & R@10 & R@50 & R@10 & R@50 & R@10 & R@50 \\
\midrule
CoSMo~\cite{lee2021cosmo}  & 23.60 & 49.18 & 18.11 & 43.18 & 24.63 & 54.31 & 22.11 & 48.89 \\
MGUR~\cite{chen2024composed}   & 23.15 & 48.74 & 18.99 & 43.47 & 25.55 & 52.83 & 22.56 & 48.35 \\
CLIP4Cir~\cite{baldrati2023composed} & 38.32 & 63.90 & 44.31 & 65.41 & 47.27 & 70.98 & 43.30 & 66.76 \\
Bi-BLIP4CIR~\cite{liu2024bi} & 39.12 & 62.92 & 39.21 & 62.81 & 44.37 & 67.06 & 40.90 & 64.26 \\
CoVR~\cite{ventura2024covr} & 44.55 & 69.03 & 48.43 & 67.42 & 52.60 & 74.31 & 48.53 & 70.25 \\
SPRC~\cite{bai2024sentencelevel} & 45.71 & 70.00 & 51.37 & 72.77 & 55.48 & 77.46 & 50.86 & 73.41 \\
QuRe~\cite{kwak2025qure} & 46.80 & 69.81 & 53.53 & 72.87 & 57.47 & 77.77 & 52.60 & 73.48 \\
\midrule
\rowcolor{gray!15} \textbf{DQE-CIR} & \textbf{48.47} & \textbf{71.09} & \textbf{55.94} & \textbf{74.62} & \textbf{59.38} & \textbf{79.12} & \textbf{54.60} & \textbf{75.94} \\
\bottomrule
\label{table1}
\end{tabular}%
}
\vspace{-4.5pt}
\end{table*}
\paragraph{FashionIQ}
In Tab.~\ref{table1}, we present a quantitative comparison of retrieval performance on the FashionIQ validation dataset, where DQE-CIR consistently surpasses existing CIR methods across categories and metrics. Using Recall@10 and Recall@50 as evaluation measures, DQE-CIR achieves the best results for each clothing type. For the Dress category, DQE-CIR attains 48.47 and 71.09, improving over the strongest prior baseline QuRe, which records 46.80 and 69.81. For the Shirt, DQE-CIR reaches 55.94 and 74.62, again outperforming QuRe with 53.53 and 72.87. A similar trend appears in the Toptee category, where DQE-CIR obtains 59.38 and 79.12, compared with 57.47 and 77.77 for QuRe. When averaged over all categories, DQE-CIR achieves 54.60 Recall@10 and 75.94 Recall@50, yielding a gain of about 2.0 to 2.5 points over the existing best method. These consistent gains under both the top-10 and the top-50 indicate that DQE-CIR produces more distinctive query embeddings that better apply the intended textual modifications, leading to more accurate retrieval of the target images across diverse garment types.

\begin{table*}[t]
\caption{Retrieval quantitative performance comparison across different CIR models on the CIRR test dataset. Results are reported using Recall@K to evaluate global ranking accuracy over the entire image corpus and Recall$_{subset}$@K to assess fine-grained target identification within the predefined candidate subset that contains the ground-truth image. The average metric jointly applies both global retrieval performance and subset-level discriminativeness.}
\vspace{3pt}
\centering
\resizebox{\textwidth}{!}{%
\begin{tabular}{lcccc|ccc|cc}
\toprule
\multirow{2}{*}{\textbf{Method}} & \multicolumn{4}{c}{\textbf{Recall@K}} & \multicolumn{3}{c}{\textbf{$\textbf{Recall}_{\textbf{subset}}\textbf{@K}$}} & \multicolumn{1}{c}{\textbf{Average}} \\
\cmidrule(lr){2-5} \cmidrule(lr){6-8} \cmidrule(lr){9-9}
 & K=1 & K=5 & K=10 & K=50 & K=1 & K=2 & K=3 & R@5 + Recall$_\textbf{subset}$@1  \\
\midrule
CoSMo & 6.48 & 23.11 & 34.63 & 67.33 & 20.29 & 40.22 & 60.80 & 43.55  \\
MGUR & 5.78 & 21.45 & 33.42 & 67.06 & 20.29 & 40.22 & 60.80 & 42.91  \\
CLIP4Cir & 44.12 & 77.23 & 86.51 & 97.95 & 73.11 & 89.11 & 95.42 & 75.17  \\
Bi-BLIP4CIR & 32.55 & 64.36 & 76.53 & 91.61 & 63.54 & 82.46 & 92.48 & 63.95  \\
CoVR & 39.76 & 70.15 & 80.89 & 95.01 & 72.46 & 87.86 & 94.77 & 71.30  \\
SPRC & 50.75 & 80.58 & 88.72 & 97.59 & 79.57 & 91.76 & 96.70 & 80.07  \\
QuRe & 52.22 & 82.53 & 90.31 & 98.17 & 78.51 & 91.28 & 96.48 & 80.52  \\
\midrule
\rowcolor{gray!15} \textbf{DQE-CIR} & \textbf{54.05} & \textbf{84.17} & \textbf{91.58} & \textbf{98.68} & \textbf{80.14} & \textbf{92.39} & \textbf{97.32} & \textbf{82.16}  \\
\bottomrule
\label{table2}
\end{tabular}%
}
\vspace{-4.5pt}
\end{table*}
\paragraph{CIRR}
The results in Tab.~\ref{table2} show the performance on the CIRR test dataset, where DQE-CIR consistently achieves the strongest retrieval performance across both global ranking accuracy and subset-level discriminativeness. Compared with prior methods, DQE-CIR attains the highest Recall@K scores at all evaluated ranks, reaching 54.05 at K=1 and 98.68 at K=50, indicating improvements in both strict top-rank retrieval and broader candidate coverage. More importantly, DQE-CIR also delivers clear gains in Recall$_{subset}$@K, achieving 80.14, 92.39, and 97.32 at K=1, 2, and 3, respectively. This demonstrates that DQE-CIR can reliably identify the correct target image within controlled candidate subsets that contain visually similar images, applying stronger fine-grained target discriminativeness under ambiguity-free retrieval settings. These improvements translate into the best overall average score of 82.16, surpassing the strongest prior baseline QuRe. The consistent gains across both global and subset-based metrics indicate that DQE-CIR learns more distinctive query representations, enabling stable and reliable target ranking even in challenging retrieval on CIRR.

 \begin{figure*}
    \centering
    \includegraphics[width=\textwidth]{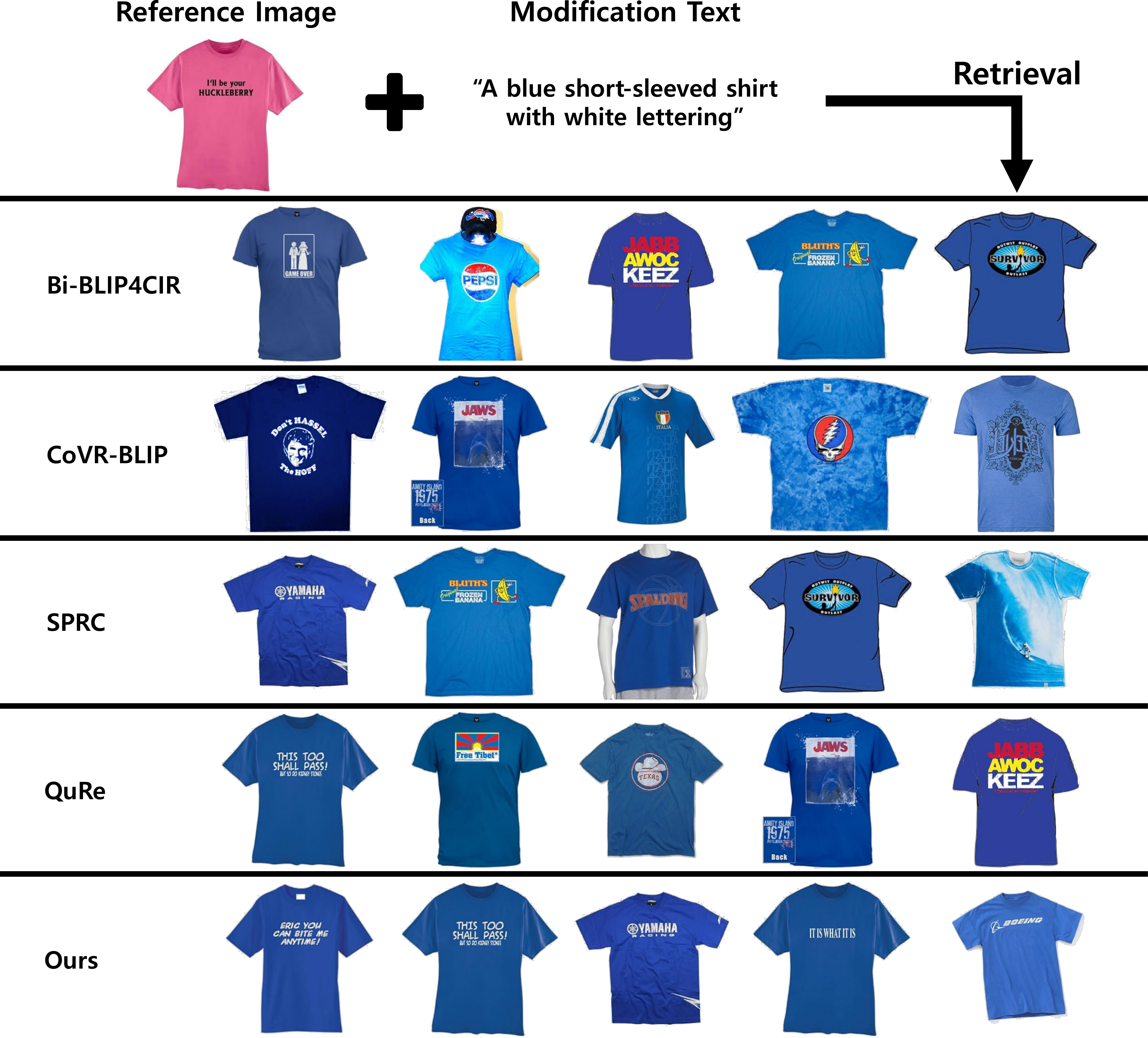}
    \caption{
    Qualitative comparison of different CIR models on the FashionIQ validation dataset.
    Given a reference image and a modification text specifying “a blue short-sleeved shirt with white lettering”, we compare the top-ranked retrieval results produced by different CIR methods. While baseline models often retrieve images that partially match the query attributes, DQE-CIR retrieves images that correctly satisfy all specified modifications.
    }
    \label{fig4}
\end{figure*}
\subsection{Qualitative Evaluation}
 We provide qualitative retrieval examples on both FashionIQ and CIRR to visually examine the behavior of CIR models under different evaluation settings. On FashionIQ, we compare DQE-CIR with other representative methods to highlight relative differences in retrieval quality and attribute alignment. On CIRR, we focus on illustrating the retrieval results of DQE-CIR under a variety of conditions, including subtle attribute changes and visually similar candidates, where accurate target identification is particularly difficult. Through these results, we illustrate how DQE-CIR produces more coherent and consistent retrieval outcomes, offering insights into the qualitative advantages that are not fully validated by numerical metrics alone.

\paragraph{FashionIQ}
The results in Fig.~\ref{fig4} shows a qualitative comparison on the FashionIQ dataset, illustrating the retrieval behavior under fine-grained attribute modification. Given a reference image of a pink short-sleeved shirt and a modification text (i.e., “A blue short-sleeved shirt with white lettering”), the goal is to retrieve images that accurately apply all specified attributes simultaneously. As shown in Fig.~\ref{fig4}, existing CIR methods tend to focus on partial feature overlap, returning top-ranked results that merely contain blue color or some white regions, regardless of whether explicit white lettering is present. This behavior indicates limited distinctive capability in capturing attribute conjunctions, as these methods prioritize coarse semantic similarity rather than precise attribute alignment. In contrast, DQE-CIR consistently retrieves blue short-sleeved shirts with clearly visible white lettering across the top results, demonstrating a more faithful interpretation of the modification text. This qualitative result highlights that the proposed method enforces stronger attribute-level reasoning in the retrieval, enabling target relative discriminativeness that filters out visually similar yet semantically incomplete candidates and yields retrievals that closely match the intended combination of color, sleeve length, and textual pattern.

\begin{figure*}
    \centering
    \includegraphics[width=\textwidth]{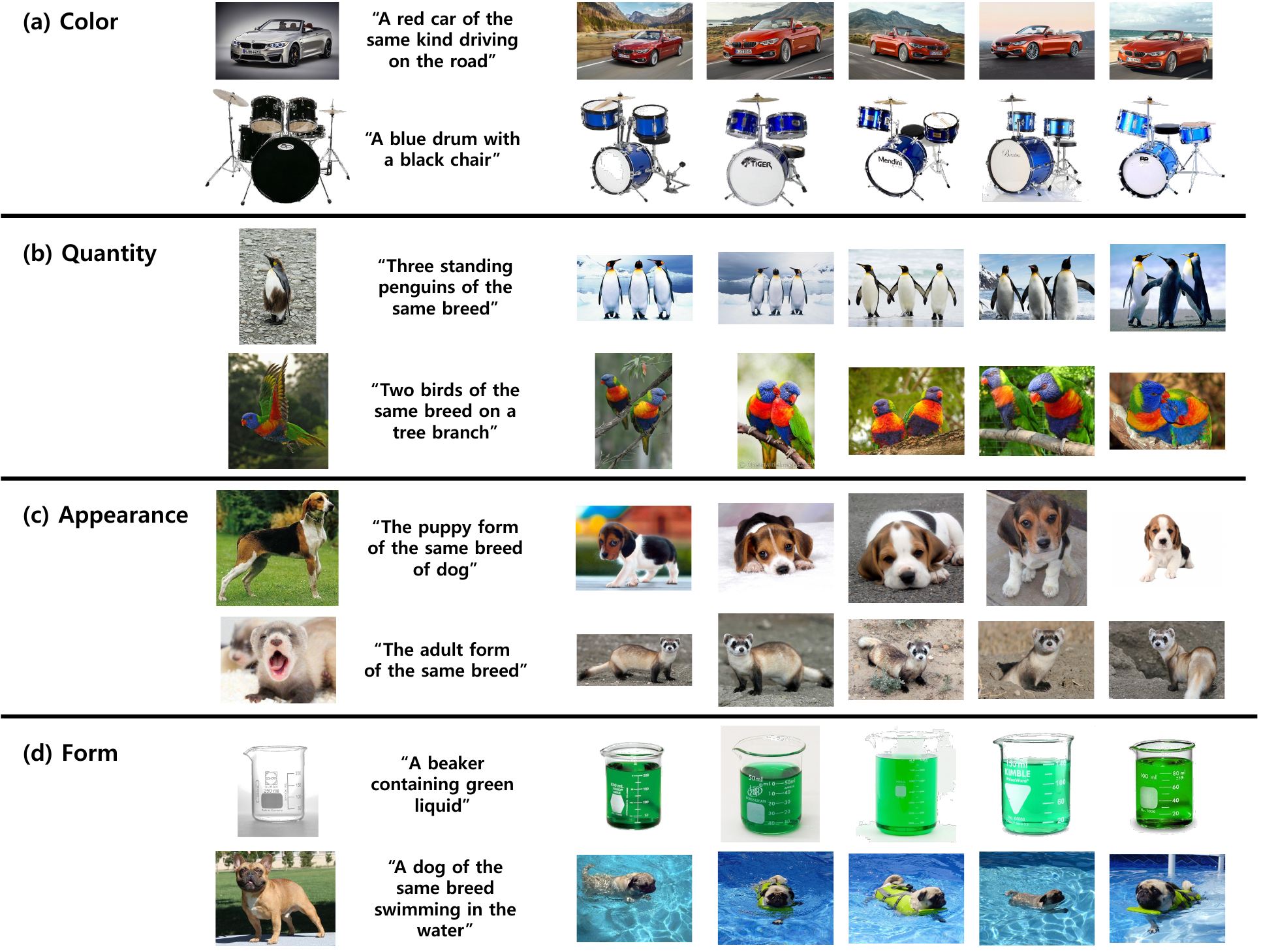}
    \caption{
    Qualitative comparison of DQE-CIR on the CIRR test dataset across different compositional categories.
    Retrieval results are conditioned on four categories of modifications: (a) Color, (b) Quantity, (c) Appearance, and (d) Form. Given a reference image and a modification text, the model retrieves images that apply the specified attribute changes while preserving the object.
    }
    \label{fig5}
\end{figure*}

\paragraph{CIRR}
The results in Fig.~\ref{fig5} provide a qualitative evaluation of the CIRR dataset, demonstrating the performance of DQE-CIR across diverse retrieval conditions driven by reference images and modification texts. As illustrated in case (a), when the modification text specifies a color change, DQE-CIR successfully retrieves images that accurately apply the transformed object color while preserving the original object identity, indicating a precise color-level feature. In case (b), the evaluation focuses on retrieving images that contain a specified number of identical objects described in the text, where DQE-CIR correctly aligns numerical constraints with the visual content and returns results that match the intended object count. Case (c) focuses on appearance changes, such as querying different life stages of the same object (i.e., younger or grown status). The retrieval results show that DQE-CIR captures these subtle appearance transitions without confusing them with semantically similar objects. Finally, case (d) examines form variations, where the modification text describes structural or contextual changes, such as background alterations or objects filled with liquid, and DQE-CIR consistently retrieves images that apply these form-level transformations. Across all cases, the qualitative results show that DQE-CIR maintains superior retrieval by jointly modeling target relative negative samples and fine-grained attribute alignment, allowing robust retrieval across variations in color, quantity, appearance, and form within a unified framework.

\begin{table}[h!]
    \centering
    \caption{Zero-shot quantitative performance comparison of diverse CIR models on the CIRCO dataset, reported using mAP, which measures the average precision computed over the ranked retrieval results.}
    \vspace{3pt}
    \resizebox{0.7\linewidth}{!}{%
    \begin{tabular}{l cccc}
        \toprule
        \textbf{Method} & \textbf{mAP@5} & \textbf{mAP@10} & \textbf{mAP@25} & \textbf{mAP@50} \\
        \midrule
        CoSMo & 0.31 & 0.40 & 0.47 & 0.53 \\
        MGUR & 0.14 & 0.17 & 0.25 & 0.30 \\
        CLIP4Cir & 10.58 & 11.18 & 12.32 & 12.96 \\
        Bi-BLIP4CIR & 4.74 & 4.97 & 5.69 & 6.10 \\
        CoVR & 18.35 & 19.25 & 21.02 & 21.88 \\
        SPRC & 17.57 & 18.48 & 20.14 & 20.98 \\
        QuRe & 23.22 & 24.23 & 26.26 & 27.24 \\
        \midrule
        \rowcolor{gray!15} \textbf{DQE-CIR} & \textbf{24.27} & \textbf{25.68} & \textbf{27.47} & \textbf{28.13} \\
        \bottomrule
    \end{tabular}}
    \vspace{-4.5pt}
    \label{table3}
\end{table}
\subsection{Ablation Studies}
We conduct ablation studies to evaluate the contribution of each component in DQE-CIR from multiple complementary perspectives. These studies are designed to systematically examine how the proposed method influences retrieval performance, representation quality, and training. By decomposing the proposed framework into its key components and analyzing their effects under controlled settings, we aim to provide a comprehensive understanding of how DQE-CIR achieves distinctive and robust CIR.


\paragraph{Zero-shot Performance Comparison}
We evaluate zero-shot retrieval performance on the CIRCO dataset~\cite{baldrati2023zero} using mAP at multiple setting levels. As shown in Tab.~\ref{table3}, DQE-CIR achieves the highest performance across all evaluation points, indicating stable generalization in the absence of dataset-specific training. DQE-CIR attains mAP@5 of 24.27, mAP@10 of 25.68, mAP@25 of 27.47, and mAP@50 of 28.13. Compared with QuRe, which reports 23.22, 24.23, 26.26, and 27.24 at the same settings, DQE-CIR shows improvements of around 1.0 to 1.5 points throughout the ranked retrieval range. Other CIR methods, including CoVR and SPRC, achieve mAP@50 values of 21.88 and 20.98, while CLIP4Cir and Bi-BLIP4CIR exhibit lower scores across all settings. Overall, the results show that DQE-CIR produces more distinctive query embeddings in zero-shot retrieval by more effectively modeling target relative negative samples and fine-grained feature alignment, leading to improved ranking quality across different retrieval depths on CIRCO.

\begin{table}[t!]
    \centering
    \fontsize{8pt}{11pt}\selectfont
    \renewcommand{\arraystretch}{1.25}
    \caption{Mid-zone range performance comparison on the CIRR test dataset. The mid-zone is defined by the $\Delta S$ interval $[\alpha,\beta]$.}
    \vspace{3pt}
    \resizebox{0.97\linewidth}{!}{%
    \begin{tabular}{lcccc|ccc}
        \toprule
        \multirow{2}{*}{\textbf{Mid-zone ($\alpha$--$\beta$)}} 
        & \multicolumn{4}{c}{\textbf{Recall@K}} 
        & \multicolumn{3}{c}{\textbf{Recall$_{\textbf{subset}}$@K}} \\
        \cmidrule(lr){2-5} \cmidrule(lr){6-8}
        & \textbf{K=1} & \textbf{K=5} & \textbf{K=10} & \textbf{K=50} 
        & \textbf{K=1} & \textbf{K=2} & \textbf{K=3} \\
        \midrule
        $0.30$--$0.70$(40\%) & 52.35 & 82.64 & 90.62 & 98.36 & 77.95 & 91.08 & 96.15 \\
        $0.25$--$0.75$(50\%) & 53.22 & 83.42 & 91.12 & 98.52 & 79.02 & 91.84 & 96.78 \\
        \rowcolor{gray!15}
        \textbf{$0.20$--$0.80$}(60\%) & \textbf{54.05} & \textbf{84.17} & \textbf{91.58} & \textbf{98.68} & \textbf{80.14} & \textbf{92.39} & \textbf{97.32} \\
        $0.15$--$0.85$(70\%) & 53.34 & 83.50 & 91.08 & 98.55 & 79.18 & 91.92 & 96.86 \\
        $0.10$--$0.90$(80\%) & 52.44 & 82.78 & 90.70 & 98.41 & 78.12 & 91.22 & 96.28 \\
        \bottomrule
    \end{tabular}%
    }
    \label{table4}
    \vspace{-4.5pt}
\end{table}

\paragraph{Performance Comparison across mid-zone Ranges}
We evaluate the effect of different mid-zone ranges defined by the $\Delta S$ interval $[\alpha, \beta]$ on the CIRR test dataset to examine how the selection of target relative negatives influences retrieval performance. As shown in Tab.~\ref{table4}, the mid-zone setting with $\alpha=0.20$ and $\beta=0.80$, corresponding to a 60\% range, yields the best overall performance across both Recall@K and Recall$_{subset}$@K metrics. Under this configuration, DQE-CIR achieves Recall@1 of 54.05, Recall@5 of 84.17, Recall@10 of 91.58, and Recall@50 of 98.68, while also obtaining Recall$_{subset}$@1 of 80.14, Recall$_{subset}$@2 of 92.39, and Recall$_{subset}$@3 of 97.32. Narrower mid-zone ranges, such as 40\% and 50\%, result in lower Recall@1 values of 52.35 and 53.22, indicating that excessively restrictive intervals reduce the availability of informative negatives. In contrast, wider ranges of 70\% and 80\% lead to decreased Recall@1 and Recall$_{subset}$@1, suggesting that overly broad intervals introduce less informative or ambiguous negatives. Overall, these results demonstrate that a balanced mid-zone range allows DQE-CIR to more effectively exploit target relative negative samples, producing more distinctive query representations and improving retrieval accuracy on CIRR.

\begin{figure*}
    \centering
    \includegraphics[width=\textwidth]{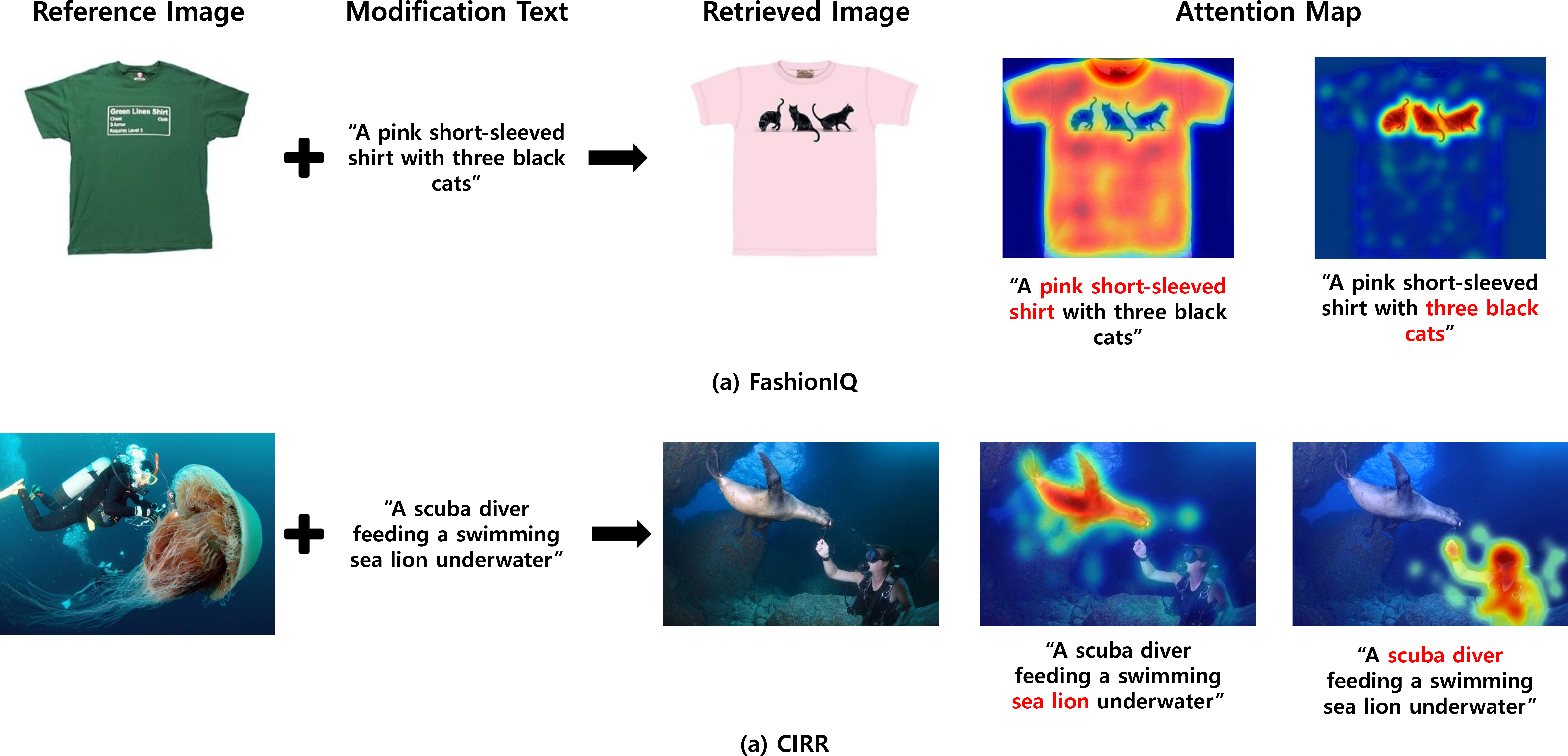}
    \caption{
    Cross-attention highlight visualization of DQE-CIR on the FashionIQ and CIRR datasets.
    The visualization highlights cross-attention maps corresponding to the composed queries, showing how the model attends to attribute-relevant regions in the retrieved images conditioned on the reference image and modification text.
    }
    \label{fig6}
\end{figure*}
\paragraph{Cross-Attention Visualization for Semantic Alignment}
We provide cross-attention visualizations to analyze how DQE-CIR achieves semantic alignment between composed queries and retrieved images. As illustrated in Fig~\ref{fig6}, the visualizations present cross-attention maps generated when the model jointly processes the reference image and the modification text, revealing which visual regions contribute most to the retrieval decision. In the FashionIQ, the attention maps consistently concentrate on attribute-specific regions corresponding to the pink color, short sleeves, and black cat patterns described in the modification text. Notably, the attention is localized around the cat silhouettes and sleeve regions rather than diffused across the entire garment, indicating precise alignment between textual attributes and their visual counterparts. In the CIRR, the attention highlights similarly concentrate on regions directly related to the target concept in the scuba diver and the sea lion. These results demonstrate that DQE-CIR emphasizes meaningful regions that directly apply the intended modification, rather than relying on entire visual appearance features. Overall, the attention visualization confirms that DQE-CIR produces semantically aligned query representations by focusing on attribute-consistent visual features, enabling precise retrieval under diverse modification conditions.

\begin{table*}[t!]
\caption{Retrieval performance comparison on the FashionIQ validation dataset with consistent vision–language model backbones.}
\vspace{3pt}
\centering
\resizebox{\textwidth}{!}{%
\begin{tabular}{llcc|cc|cc|cc}
\toprule
\multirow{2}{*}{\textbf{Method}} &
\multirow{2}{*}{\textbf{Backbone}} &
\multicolumn{2}{c}{\textbf{Dress}} & \multicolumn{2}{c}{\textbf{Shirt}} & \multicolumn{2}{c}{\textbf{Toptee}} & \multicolumn{2}{c}{\textbf{Average}} \\
\cmidrule(lr){3-4} \cmidrule(lr){5-6} \cmidrule(lr){7-8} \cmidrule(lr){9-10}
 & & R@10 & R@50 & R@10 & R@50 & R@10 & R@50 & R@10 & R@50  \\
\midrule
Bi-BLIP4CIR & BLIP & 39.12 & 62.92 & 39.21 & 62.81 & 44.37 & 67.06 & 40.90 & 64.26 \\
QuRe & BLIP & 40.80 & 64.90 & 45.93 & 65.90 & 52.07 & 72.87 & 46.27 & 67.89 \\
\rowcolor{gray!15}
\textbf{DQE-CIR} & \textbf{BLIP} & \textbf{41.87} & \textbf{65.92} & \textbf{47.16} & \textbf{66.83} & \textbf{53.58} & \textbf{74.24} & \textbf{47.54} & \textbf{69.03} \\
\midrule
CoVR-2 & BLIP-2 & 46.41 & 69.51 & 49.75 & 67.76 & 51.86 & 72.46 & 49.34 & 69.91 \\
QuRe & BLIP-2 & 46.80 & 69.81 & 53.53 & 72.87 & 57.47 & 77.77 & 52.60 & 73.48 \\
\rowcolor{gray!15}
\textbf{DQE-CIR} & \textbf{BLIP-2} & \textbf{47.83} & \textbf{70.68} & \textbf{54.89} & \textbf{73.94} & \textbf{59.02} & \textbf{79.11} & \textbf{53.91} & \textbf{74.55} \\
\bottomrule
\label{table5}
\end{tabular}
}
\vspace{-4.5pt}
\end{table*}
\paragraph{Performance Comparison with Consistent Backbones}
We report a controlled quantitative comparison on the FashionIQ validation dataset using the same vision-language backbone. As shown in Tab.~\ref{table5}, when the BLIP backbone is fixed, DQE-CIR consistently outperforms existing baselines Bi-BLIP4CIR and QuRe across all categories and both Recall@10 and Recall@50. Compared with the strongest prior baseline QuRe under the same BLIP setting, DQE-CIR improves Dress from 40.80 and 64.90 to 41.87 and 65.92, Shirt from 45.93 and 65.90 to 47.16 and 66.83, and Toptee from 52.07 and 72.87 to 53.58 and 74.24, resulting in a higher average from 46.27 and 67.89 to 47.54 and 69.03. Since all methods share an identical backbone, these gains apply to the contribution of the proposed method and query construction strategy rather than architectural differences. Consistent improvement is also observed under the stronger BLIP-2 backbone, where DQE-CIR again achieves the best performance in every category, surpassing QuRe from 46.80 and 69.81 to 47.83 and 70.68 on Dress, from 53.53 and 72.87 to 54.89 and 73.94 on Shirt, and from 57.47 and 77.77 to 59.02 and 79.11 on Toptee, yielding an average improvement from 52.60 and 73.48 to 53.91 and 74.55. These consistent improvements under fixed backbones demonstrate that the proposed target relative negative sampling and pairwise learning enable the construction of more distinctive query embeddings and accurate retrieval of fine-grained attribute modifications, validating their effectiveness independently of the backbone.

\begin{figure*}
    \centering
    \includegraphics[width=\textwidth]{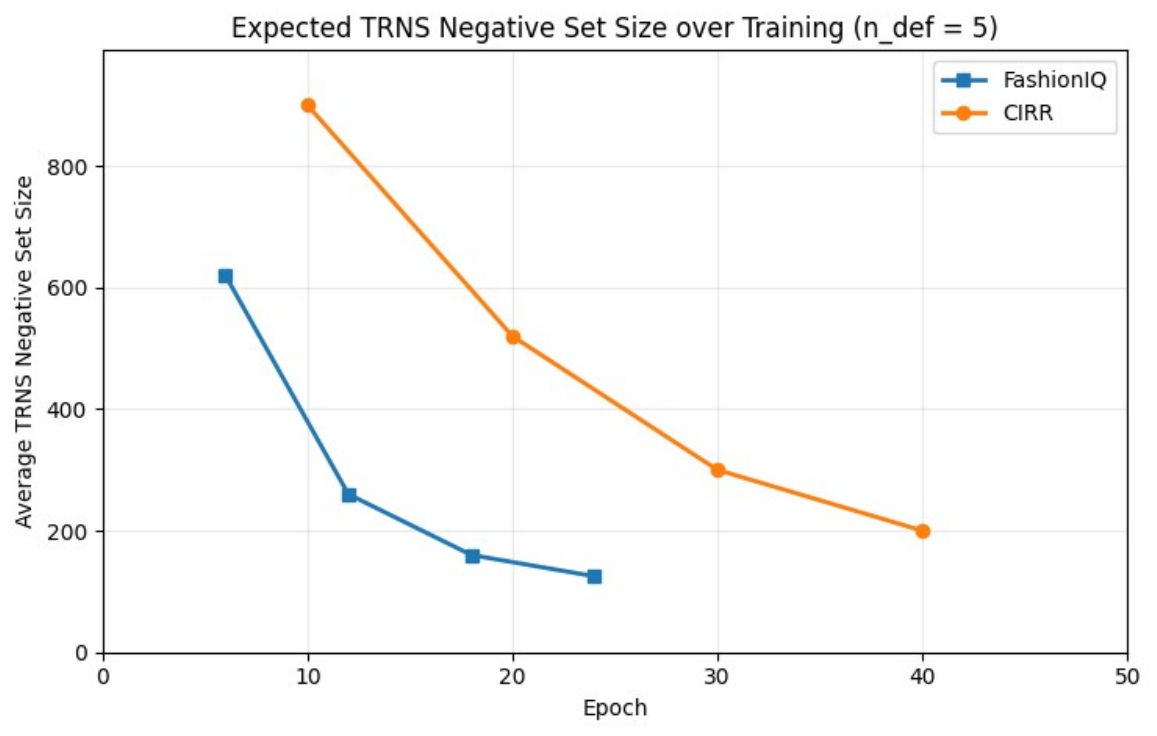}
    \caption{
        Average TRNS negative set size across training on FashionIQ and CIRR. 
        The negative set is redefined based on the target relative similarity distribution. The decreasing trend indicates that the distribution becomes progressively sharper, allowing the method to focus on a more informative set of negatives as training proceeds.
    }
    \label{fig7}
\end{figure*}
\paragraph{TRNS Negative Set Size over Training}
We analyze the evolution of the average TRNS negative set size across training epochs on FashionIQ and CIRR to better understand the behavior of the proposed target relative negative sampling. As illustrated in Fig.~\ref{fig7}, the average number of target relative negatives per query at the epochs where the negative set is redefined. For both datasets, the average set size is initially large at the first redefinition stage, indicating that many candidate images exhibit similar relevance scores to the target when the embedding space is still coarse. As training progresses, the average TRNS negative set size decreases consistently across successive redefinition epochs. This trend suggests that the learned embedding space becomes increasingly structured, allowing the method to more effectively separate the target from non-target candidates based on target relative similarity differences. Notably, the reduction is more rapid on FashionIQ, applying its relatively constrained visual domain, while CIRR exhibits a more gradual decline due to its higher visual diversity and more complex retrieval conditions. Overall, this behavior confirms that DQE-CIR progressively sharpens target relative similarity distributions during training, enabling the method to focus on a smaller and more informative set of negatives, which supports more stable optimization and more distinctive query embeddings in CIR.

\section{Conclusion}
In this paper, we propose DQE-CIR, a method designed to construct more distinctive and attribute-aware query embeddings by jointly addressing relevance suppression and semantic confusion inherent in existing contrastive learning frameworks. DQE-CIR introduces learnable attribute weights that explicitly control the contribution of color- and shape-sensitive features within the composed query representation, enabling the model to faithfully apply fine-grained modification intents while maintaining clear separation from visually similar distractors. In addition, we formulate target relative negative sampling, which defines a $\Delta S$-based mid-zone to exclude false negatives and overly easy negatives, allowing training to focus on semantically informative samples that provide effective supervision. By combining this target relative sampling strategy with single-negative pairwise ranking, DQE-CIR enforces a clear preference structure between the target image and carefully selected negatives, leading to a more stable and distinctive embedding space. Extensive experiments on FashionIQ, CIRR, and CIRCO demonstrate that DQE-CIR consistently outperforms existing methods across both global retrieval and subset-level evaluation metrics, with particularly strong gains under fine-grained and ambiguous retrieval conditions. Qualitative results show that, compared to existing methods, DQE-CIR retrieves images that more faithfully satisfy fine-grained modification intents, particularly in visually similar cases. In ablation studies, we further confirm that each component of DQE-CIR contributes meaningfully to the observed performance gains, validating the effectiveness of learnable attribute weighting, target relative negative sampling, and single-negative pairwise learning in constructing distinctive query embeddings. Overall, the proposed method provides an effective and unified solution for attribute-sensitive image retrieval, offering improved retrieval accuracy across retrieval tasks.

\section{Acknowledgment}
This research was supported by the Institute of Information \& Communications Technology Planning \& Evaluation (IITP) grant, funded by the Korea government (MSIT) (No. RS-2019-II190079 (Artificial Intelligence Graduate School Program (Korea University)), No. IITP-2025-RS-2024-00436857 (Information Technology Research Center (ITRC), and No.IITP-2025-RS-2025-02304828 (the artificial intelligence star fellowship support program to nurture the best talents).


\clearpage
\bibliographystyle{elsarticle-num} 
\bibliography{ref}






\end{document}